\documentclass{article}

\usepackage{arxiv}

\usepackage[utf8]{inputenc} 
\usepackage[T1]{fontenc}    
\usepackage{hyperref}       
\usepackage{url}            
\usepackage{booktabs}       
\usepackage{amsfonts}       
\usepackage{nicefrac}       
\usepackage{microtype}      
\usepackage{lipsum}
\usepackage[pdftex]{graphicx}
\usepackage{xcolor}

\usepackage{amsmath,xparse}

\title{Hierarchical Action Classification with Network Pruning}

\author{
  Mahdi Davoodikakhki\\
  School of Computing Science\\
  Simon Fraser University\\
  \texttt{mahdid@sfu.ca}\\
   \And
 KangKang Yin \\
  School of Computing Science\\
  Simon Fraser University\\
  \texttt{kkyin@sfu.ca}\\
}

\definecolor{kkBlue}{rgb}{0,0.0,0.9}

\definecolor{maRed}{rgb}{0.9,0.0,0.0}

\begin{document}

\maketitle

\begin{abstract}
Research on human action classification has made significant progresses in the past few years. Most deep learning methods focus on improving performance by adding more network components. We propose, however, to better utilize auxiliary mechanisms, including hierarchical classification, network pruning, and skeleton-based preprocessing, to boost the model robustness and performance. We test the effectiveness of our method on four commonly used testing datasets: NTU RGB+D 60, NTU RGB+D 120, Northwestern-UCLA Multiview Action 3D, and UTD Multimodal Human Action Dataset. Our experiments show that our method can achieve either comparable or better performance on all four datasets. In particular, our method sets up a new baseline for NTU 120, the largest dataset among the four. We also analyze our method with extensive comparisons and ablation studies.
\end{abstract}


\section{Introduction}
{Human action classification and recognition has many important applications, such as autonomous driving, smart surveillance, patient monitoring, and interactive games. Despite extensive research on this topic in recent years, human-level performance is still out of reach. Image classification, however, has achieved human-level performance a few years ago. There are many challenges in human action recognition. First, there are high intra-class variations and inter-class similarities. A powerful deep learning model and a large amount of training data are necessary to achieve good performance. Second, the qualities of input videos vary greatly. There are multiple benchmark datasets, and in this work we focus on captured indoor videos in lab environments. Third, multiple data types and representations can be captured with the video data or extracted from the videos. Skeleton data, for example, should be used whenever possible.}

{We propose to extend the Inflated ResNet architecture with hierarchical classification for better feature learning at different scales. Iterative pruning is then incorporated for a further performance boost. We also use skeleton data, captured or extracted, to crop out irrelevant background so the learning can focus on human activities. These mechanisms combined set up a new baseline for the newly-released large-scale dataset NTU RGB+D 120.}

{In summary, our main contributions include:
\begin{itemize}
  \item We show that Inflated ResNet coupled with hierarchical classification can boost the performance of the baseline model. 
  \item We show that iterative pruning can help improve the performance even further.
  \item We also show that 2D/3D skeleton data, when available, could be used to crop videos in a preprocessing stage to increase the classification accuracy in most cases.
  \item We evaluate our method extensively on four datasets. Our method sets up a new baseline for the NTU RGB+D 120 dataset for future research in this field.
\end{itemize}}

\section{Related Work}
{There is a large body of prior work related to our work. Due to the limited space, we only summarize the most relevant and most recent papers here.}

\subsection{Human Action Classification}

{Most human action classification methods work on either RGB image sequences and/or skeleton data. Our method uses both 2D skeletons and video inputs for classification, so we will review and compare with state-of-the-art methods from both categories. However, we only use skeletons for preprocessing to crop out irrelevant backgrounds, and not for classification.} 
 
\subsubsection{Skeleton-based action classification}

{For skeleton-based classification, traditional CNN (Convolutional Neural Networks) methods can still be used after converting skeleton data into 2D images. Example works include TSRJI~\cite{caetano2019skeleton}, Skelemotion~\cite{Caetano_2019}, Enhanced Viz.~\cite{articleLiu}, JTM~\cite{Wang_2016}, JDM~\cite{Li2017Joint}, and Optical Spectra~\cite{Hou2018Skeleton}. RNN (Recurrent Neural Networks) and its two common variations LSTM (Long Short-Term Memories) and GRU (Gated Recurrent Units) can also be used to interpret skeleton sequences. Their ability to learn long and short-term memories help achieve good results. Example works include TS-LSTM~\cite{Lee2017enesmble} and EleAtt-GRU~\cite{zhang2018adding}. Most advanced methods, however, are based on Graph Convolutional Networks (GCN), which can model sparse joint connections. Example works include MS-G3D Net~\cite{liu2020disentangling}, DGNN~\cite{Shi_2019_Skeleton}, 2s-AGCN~\cite{shi2018twostream}, FGCN~\cite{yang2020feedback}, and GVFE + AS-GCN with DH-TCN ~\cite{papadopoulos2019vertex}. } 

\subsubsection{Video-based action classification}
{Most state-of-the-art video-based classification methods are based on CNN. Inflated 3D ConvNet (I3D) and Inflated ResNet proposed by \cite{carreira2017quo} became the foundation of many advanced algorithms, such as MMTM~\cite{joze2019mmtm}, Glimpse Clouds~\cite{Baradel_2018}, Action Machine~\cite{zhu2018action}, and PGCN~\cite{shi2019action}. Such networks inflate 2D kernels of convolutional filters and pooling layers along the temporal domain to process 3D spatio-temporal information. In addition, Glimpse Clouds~\cite{Baradel_2018} extracts attention glimpses from each frame and uses the penultimate feature maps to estimate 2D joint positions and encourage glimpses to focus on the people in the scene. Action Machine ~\cite{zhu2018action} extracts Region of Interests (RoI), which are human bounding boxes, for better pose estimation and classification over these regions. The skeleton classification results are then fused with video-based classification to boost the performance further. PGCN~\cite{shi2019action} performs graph convolutions over RGB features extracted around 2D joints rather than over the joint positions to improve the video-based classification performance, which is then also fused with skeleton-based classification scores. Our baseline network is similar to that of Glimpse Clouds~\cite{Baradel_2018}, and our cropping preprocess is inspired by Action Machine ~\cite{zhu2018action}.}

{In addition to video and skeleton input, various other data types can be used for input or intermediate feature representations. For instance,  PoseMap~\cite{Liu2018Recognizing} extracts pose estimation maps from the RGB frames. 3D optical flow can also be estimated for classification~\cite{Ballin_3DFlowEstimation}. RGB and optical flow classifications can be fused to further boost performance~\cite{Simonyan_two-stream}. MMTM~\cite{joze2019mmtm} uses a combination of depth, optical flow, and skeleton data with RGB frames as input for different datasets. Different fusion strategies have also been investigated, such as MMTM~\cite{joze2019mmtm} and \cite{Perez_Rua_2019} that fuse features from intermediate layers for the next layers or final classification.}

\subsection{Hierarchical Classification and Loss Functions}
{Hierarchical classification and loss functions facilitate learning the most important features in different scales. One straightforward way to apply hierarchical classification is to learn on different resolutions of the same set of images, such as \cite{Kawahara2016MultiresolutionTractCW} for skin lesion classification. Semantic graphs can be constructed to form hierarchical relations among classes for text classification~\cite{Wu_2019}. Our approach is mainly inspired by related works in image classification that use features from intermediate layers for the main classification as well, by accumulating loss functions from all participating layers~\cite{Szegedy_2015_CVPR,huang2018multiscale}.}

\subsection{Network Pruning}
{Over-parameterization is a well-known property of deep neural networks. Network pruning is usually used to improve generalization and achieve more compact models for low-resource applications ~\cite{frankle2018lottery,liu2018rethinking}. There are multiple choices to implement pruning and the fine-tuning after pruning. One option is to use one-shot pruning \cite{li2016pruning}, but usually unimportant filters in convolutional layers are iteratively located and deleted. After each pruning iteration, the network can be re-trained from scratch with reinitialized weights~\cite{frankle2018lottery,liu2018rethinking}. Our pruning method is similar to~\cite{frankle2018lottery}, as we also iteratively prune filters with the lowest $l2$ norms. The difference is that we retrain with inherited weights, similar to~\cite{li2016pruning}.}

\section{Our Methods}

{We present our video preprocessing procedures including cropping and projecting 3D skeletons to 2D in Section~\ref{sec:preprocessing}. We describe our modified ResNet network architecture in Section~\ref{sec:network}. We then detail the hierarchical classification and network pruning in Section~\ref{sec:hierarchy} and \ref{sec:prunning}, respectively.}

\subsection{Video Preprocessing}
\label{sec:preprocessing}
{Raw action videos usually contain not only human subjects, but also surrounding objects and background environments, most of which are irrelevant to the performed actions. Neural networks can overfit to such noises instead of focusing on human actions. Inspired by \cite{zhu2018action}, we also crop the raw videos to focus on human skeletons inside.}

{We use 2D skeletons and joint positions in pixels for cropping. 2D skeleton data can be captured together with the videos, or extracted by pose estimation algorithms such as OpenPose~\cite{OpenPose}, or computed from 3D skeletons as we will explain shortly. We first extract the skeleton bounding boxes, and then enlarge them by a factor of 10\% on all four sides and cap them at frame borders, in order to leave a margin for errors and retain relevant information of surrounding areas. After cropping, we rescale all the video frames to a resolution of $256 \times 256$ as input to our neural network.}

\begin{figure}[tb]
\centering
\includegraphics[width=\linewidth]{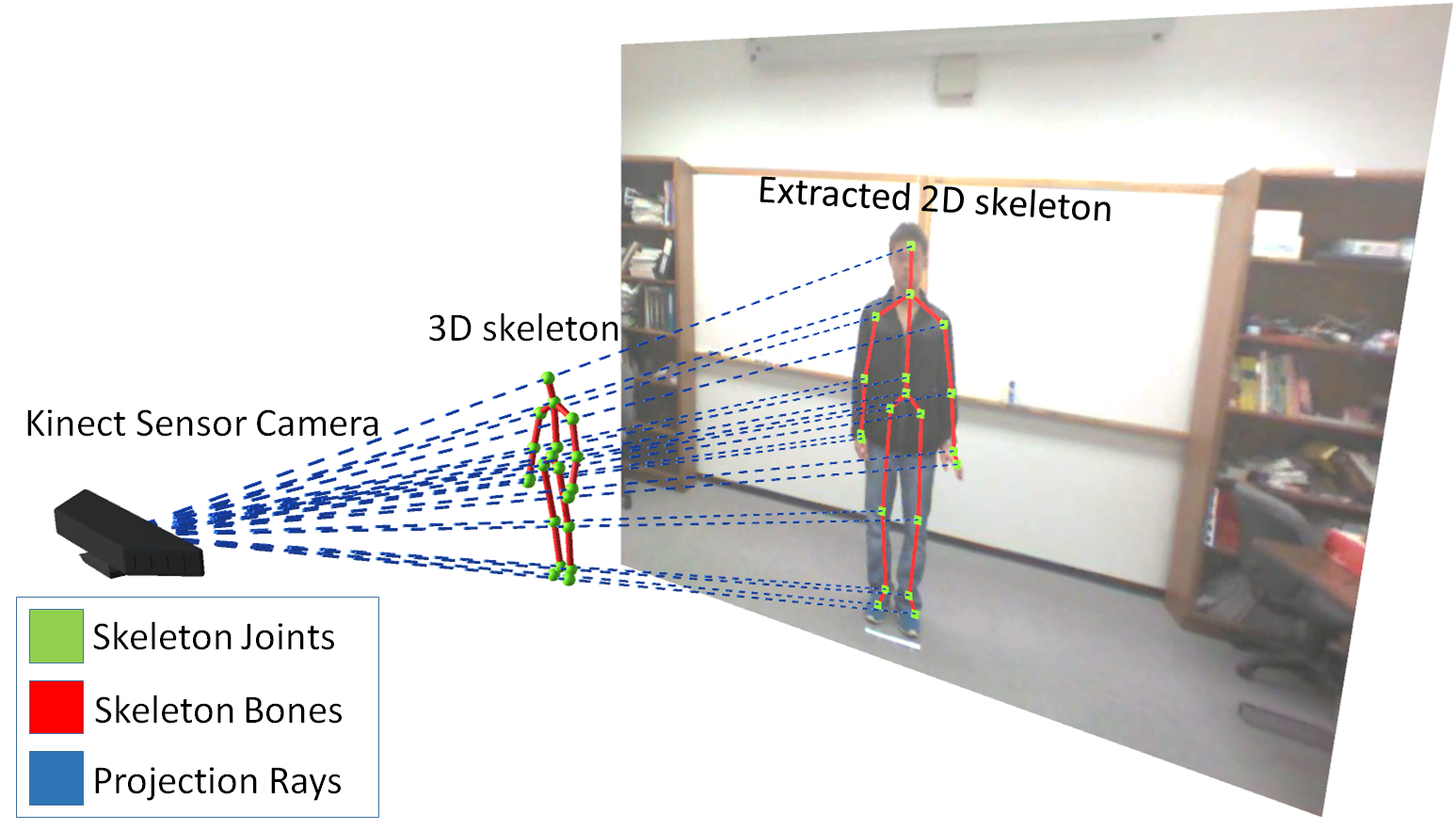}
\caption{Projection of a 3D skeleton onto the image plane as a 2D skeleton.}
\label{fig:projection}
\end{figure}

{For datasets that provide 3D skeleton data, we project the 3D skeletons onto the image plane as 2D skeletons as illustrated in Figure~ \ref{fig:projection} following Equation~\ref{projection}. We denote the 2D and 3D skeletons as $S_p \in R ^ {T \times J \times 2}$ and $S \in R ^ {T \times J \times 3}$, respectively, where $T$ is the number of frames and $J$ is the number of joints. We denote the individual channels in skeleton data as $S_{px}, S_{py}$ for 2D pixel positions, and $S_x, S_y, S_z$ for 3D world coordinates. $b_x=320$ and $b_y=240$ are bias values that correspond to the image centers for both the N-UCLA and UTD-MHAD datasets. $c_x,c_y$ are coefficients that can be solved for from Equation~\ref{calculation}. We randomly sample ten frames from UTD-MHAD video clips and manually estimate the pixel position $S_{px}$ and $S_{py}$ of 5 end-effector joints (head, hands, and feet). A least squares fit returns $c_x=558.1$ and $c_y=579.5$. These coefficients work well for both the N-UCLA and UTD-MHAD datasets.} 

\begin{equation} 
\label{projection}
S_p = \begin{bmatrix}
S_{px}\\
S_{py}
\end{bmatrix} =
\begin{bmatrix}
c_x \times\frac{S_x}{S_z} + b_x \\
c_y \times\frac{S_y}{S_z} + b_y
\end{bmatrix}
\end{equation}
 
\begin{equation} 
\label{calculation}
\begin{bmatrix}
c_x\\
c_y
\end{bmatrix} =
\begin{bmatrix}
(S_{px} - b_x)  \times \frac{S_z}{S_x} \\
(S_{py} - b_y)  \times \frac{S_z}{S_y}
\end{bmatrix}
\end{equation}

\subsection{Modified ResNet Architecture}
\label{sec:network}

{Our baseline network is the Inflated ResNet from Glimpse Clouds~\cite{Baradel_2018}, which was created according to the inflation procedure introduced in \cite{carreira2017quo}. The Inflated ResNet is a variation of ResNet developed by \cite{He_2016}. It is also similar to the I3D network in \cite{zhu2018action}. In Inflated ResNet, 2D convolutional kernels, except the first one, are converted into 3D kernels, to make them suitable for video input. We perform experiments on different variations of Inflated ResNet in Section~\ref{sec:resnet}, and find the Inflated ResNet50 to be the best architecture for our task.} 



\begin{figure}[tb]
\centering
\includegraphics[width=\linewidth]{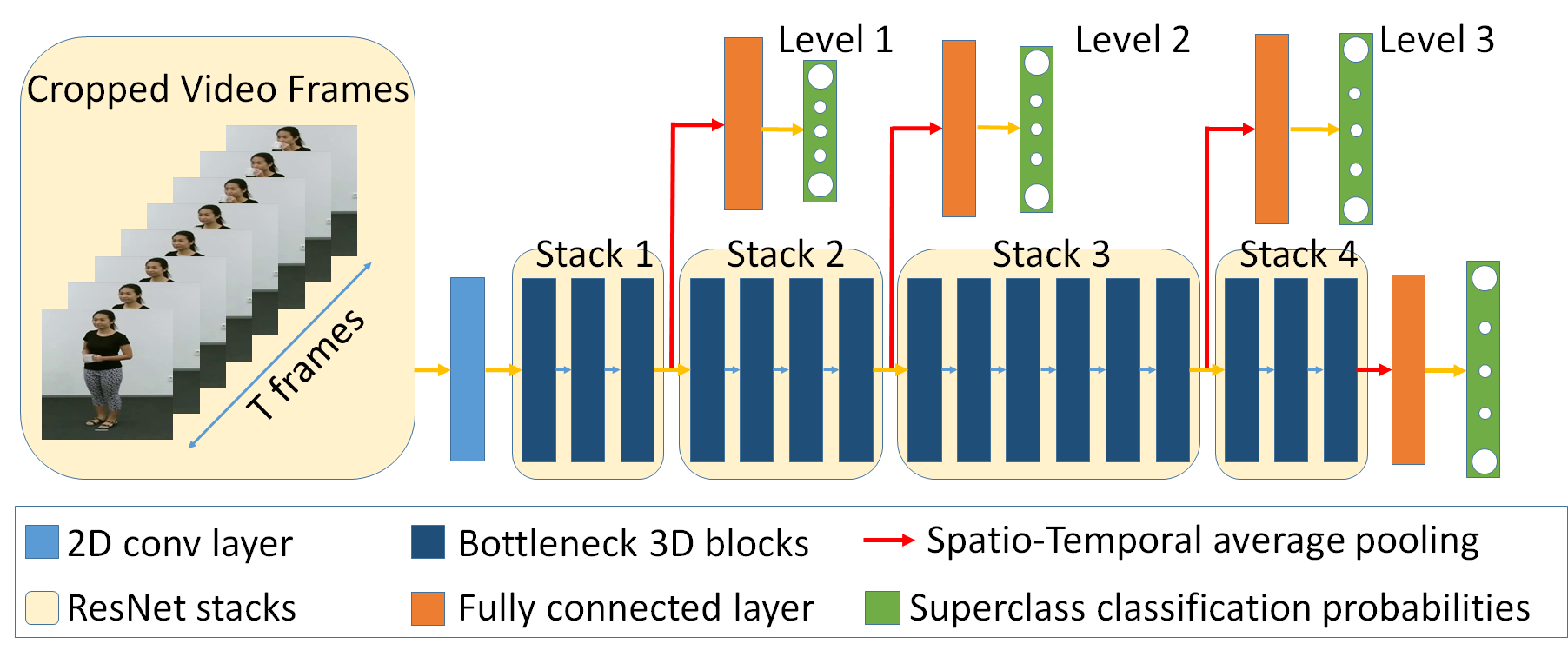}
\caption{Structure of our neural network.}
\label{fig:network}
\end{figure}

{Our baseline network consists of four main ResNet stacks of convolutional layers, each consisting of multiple bottleneck building blocks. We label them as stacks 1 to 4 in Figure~\ref{fig:network} for hierarchical classification. More specifically, we modify the baseline network after each ResNet stack by averaging the extracted features over the spatial and temporal domain, and then passing them to a fully-connected linear layer and a softmax layer to obtain multiple levels of classification probabilities.} 


\subsection{Hierarchical Classification}
\label{sec:hierarchy}
{We employ hierarchical classification to encourage our neural network to learn important features at different scales. For each ResNet stack, we enforce a superclass constraint for each action. That is, a fully-connected linear classifier assigns each action to a superclass after each ResNet stack. Each superclass contains the same number of original action classes to keep the learned hierarchy balanced. We use up to four ResNet stacks for hierarchical classification. The final structure of our network is shown in Figure~\ref{fig:network}. We train the classifiers in two passes. The first pass trains the network with a cross-entropy loss only after the last stack. We then compute the confusion matrix $C$ from the trained network for assigning the superclasses as described next. }


{We define a graph with $N$ nodes, each corresponding to one of the original action classes. Edge $e_{ij}$ denotes a connection between node $i$ and $j$ ($i < j$), with its cost defined as $e_{ij} = c_{ij} + c_{ji}$, where $c_{ij}$ is the corresponding element of the confusion matrix $C$. We denote the assigned superclass for each action class as $s_{i}^{l} \in \{1,...,N/M_l\}$, where $i \in \{1,...,N\}$ is the action class index, $l \in \{1,...,L\}$ is the stack index, and hyperparameter $M_l$ is the number of superclasses of stack $l$. The bigger $l$ is, the larger $M_l$ is. That is, front stacks have fewer number of superclasses which only need to find the most distinguishable features to classify the actions. The rear stacks, however, need to concentrate more on finer details to differentiate actions into more categories. We aim to minimize the sum of edge costs among all superclasses at all levels:
\begin{equation}\label{cost function}
     \min \sum_{i,j}{e_{ij}}, \hspace{10pt} \forall \hspace{1pt} i,j \in \{1,...,N\}, s_{i}^{l} \neq s_{j}^{l}, l \in \{1,...,L\} \quad
\end{equation}
}

\begin{figure}[tbh]
\centering
\includegraphics[width=\linewidth]{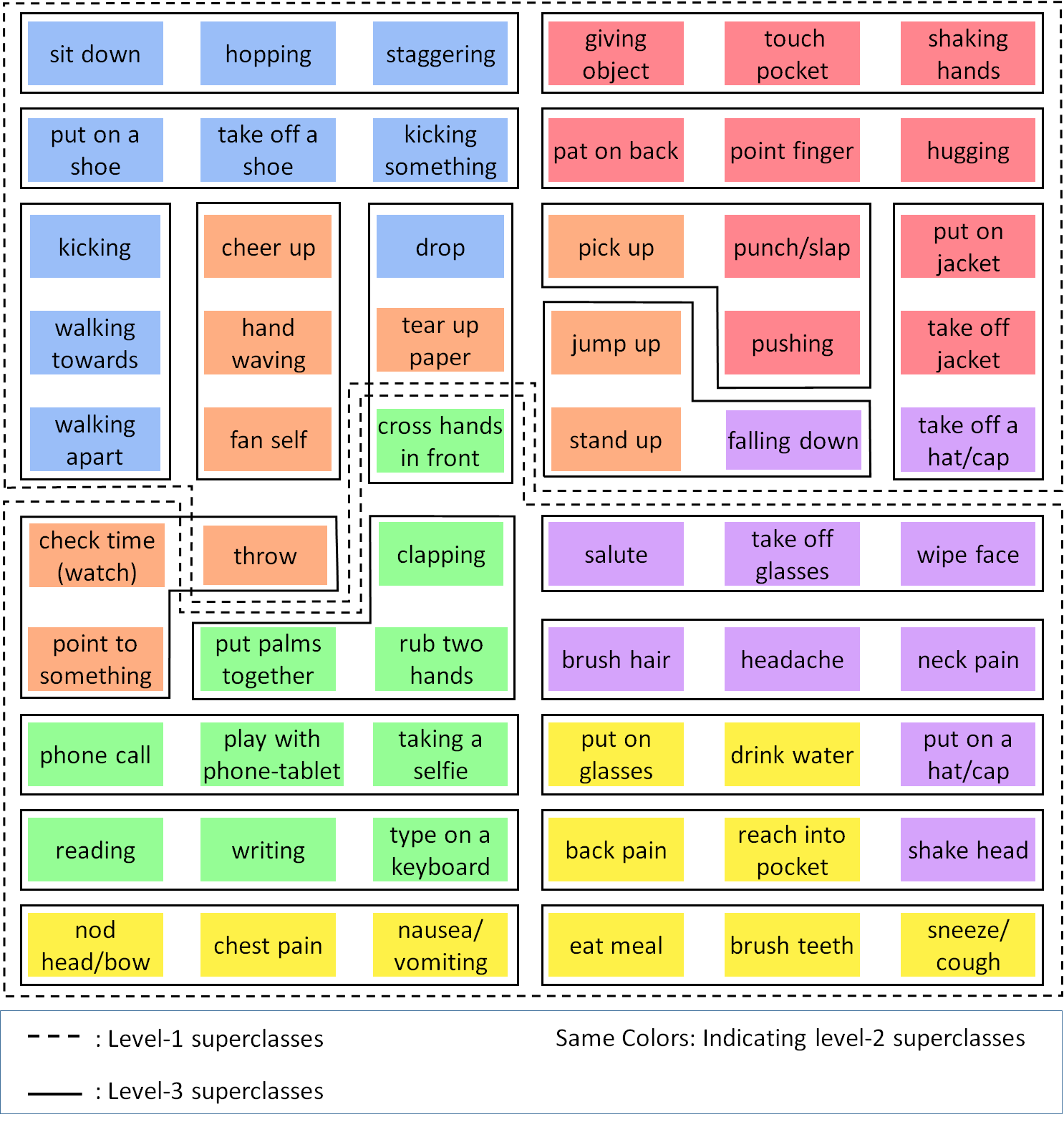}
\caption{NTU 60: all 60 action classes and the derived superclasses. }
\label{fig:ntu_classes}
\end{figure}

\begin{figure}[tbh]
\centering
\includegraphics[width=\linewidth]{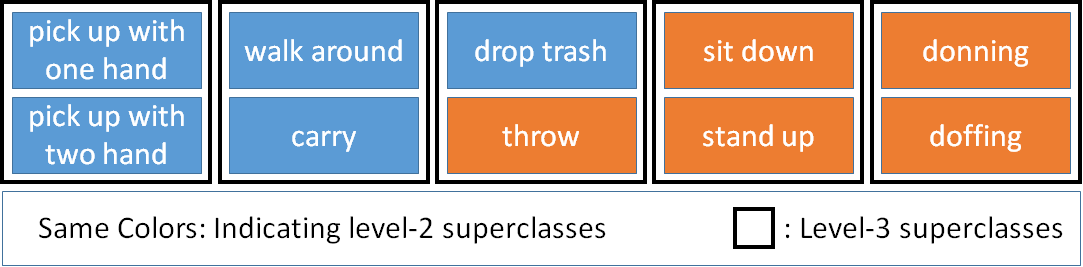}
\caption{N-UCLA: all 10 action classes and the derived superclasses.}
\label{fig:nucla_classes}
\end{figure}

{We use a simple greedy algorithm to minimize the total edge cost. We initialize the superclass assignments randomly but evenly. At each optimization step we swap two superclass assignments that decrease the cost function the most. We continue this procedure until no more deduction can be achieved. We run the greedy optimization 1000 times and then choose the solution with the lowest cost. More advanced optimization algorithms, such as genetic algorithms or deep learning methods, can be used as well. But our early experiments showed similar performance among different algorithms. So we choose the simple greedy algorithm in the end.}

{The optimized superclass assignments help classify the original action classes into similar groups. Figure~\ref{fig:ntu_classes} shows one example for the NTU 60 dataset. All mutual classes with two persons in action are classified into the same level-1 superclass. Most medical actions are in the other superclass. Figure~\ref{fig:nucla_classes} shows another example for the N-UCLA dataset. All class pairs in the level-3 superclass are similar. For instance the ``carry'' and ``walk around'' classes or the ``throw'' and ``drop trash'' classes.}

\begin{figure}[tbh]
\centering
\includegraphics[width=\linewidth]{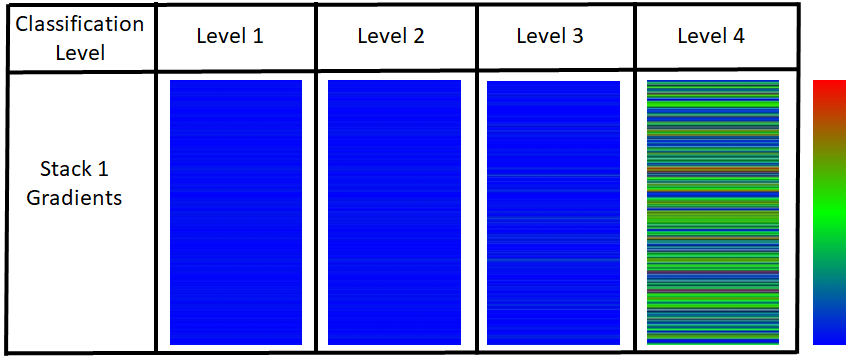}
\caption{Averaged and normalized absolute gradients in the last convolutional layer of stack 1, backpropagated from different levels of classifiers. The gradients are computed from the first 500 clips of the NTU 60 dataset. Pure red indicates 1 and pure blue indicates 0.}
\label{fig:level-1gradients}
\end{figure}

\subsubsection{Hierarchical Loss Function}
{After the first pass of training, we denote the predicted classification probabilities by each ResNet stack as $\tilde{y}_l$. We also denote the ground truth and the superclass assignments from the greedy algorithm as $y_l$. We then train the network for a second time, using a cross-entropy loss after each stack denoted as $Loss_{l}(y_{l}, \tilde{y}_l)$. The total loss is a weighted sum of the loss for each stack:
\begin{equation}\label{loss}
    Loss(y_{l}, \tilde{y}_l) = \sum_{l}w_l \times Loss_{l}(y_{l}, \tilde{y}_l), \hspace{5pt} l \in \{1,...,L\}
\end{equation}
Generally speaking, we use bigger weights $w_l$ for latter stacks. This is because the prior stacks tend to extract low-level features less important for the final classification. In Figure~\ref{fig:level-1gradients} we show the normalized and averaged absolute gradients backpropagated from each classifier to the last convolutional layer of stack 1, when using equal weights $(1,1,1,1)$ for all ResNet stacks. Note that gradients from latter stacks are richer than those of the previous stacks. We also experiment with different weighting schemes in Table~\ref{tbl:loss_weigths_comparison}.}

\subsection{Network Pruning}
\label{sec:prunning}

{We apply network pruning to further improve the accuracy and generalization ability of our model. We also tried to apply dropout but did not find it beneficial in our experiments. We employ two iterative pruning methods while keeping the hierarchical classification approach intact. } 

{For both pruning methods, we find the $p\%$ of the convolutional filters with the lowest $l2$ norm and zero them out. The first method prunes the bottom $p\%$ of filters in all convolutional layers of the ResNet stacks; while the second method prunes the bottom $p\%$ of filters in each layer of the ResNet stacks independently. We compare different choices of $p$ values on the NTU 60 dataset using the first pruning method in Table~\ref{tbl:pruning_comparison}. We will use $10\%$ for all our experiments in Section~\ref{sec:experiments}. The pruning can be carried out iteratively for multiple passes. We perform pruning until we observe reductions in testing performance for two consecutive steps. We will show more pruning results for up to seven passes in Tables \ref{tbl:NTU_60_ab}-\ref{tbl:UTD_ab} in our ablation studies.}

\begin{table}[tbh]
\begin{center}
\caption{Different weighting schemes of the hierarchical loss function on the NTU 60 dataset.}
\label{tbl:loss_weigths_comparison}
\begin{tabular}{|c|c|c|c|}
\hline
Superclass weights & Cross-Subject & Cross-View\\ 
\hline
$\frac{1}{64}, \frac{1}{16}, \frac{1}{4}, 1$  & 95.36\% & 98.29\%\\ 
$\frac{1}{27}, \frac{1}{9}, \frac{1}{3}, 1$  & 95.07\% & 98.41\% \\ 
$\frac{1}{8}, \frac{1}{4}, \frac{1}{2}, 1$  & \textbf{95.45}\% & \textbf{98.59}\% \\ 
$1, 1, 1, 1$  & 95.31\% & 98.01\% \\ 
$1, \frac{1}{2},\frac{1}{4},\frac{1}{8}$ & 93.91\%  & 97.20\% \\ 
$1, \frac{1}{3},\frac{1}{9},\frac{1}{27}$ & 93.22\%  & 97.24\% \\ 
$1, \frac{1}{4}, \frac{1}{16}, \frac{1}{64}$  & 92.21\% & 95.85\%\\ 

\hline

\end{tabular}
\end{center}
\end{table}

\begin{table}
\begin{center}
\caption{Different pruning ratios and passes on the NTU 60 cross-subject benchmark.}
\label{tbl:pruning_comparison}
\begin{tabular}{|c|c|c|c|}
\hline
Pruning passes & 5\% pruning &  10\% pruning &  15\% pruning \\
\hline
Pass 1 pruning & 94.84\% & 95.52\% & 95.08\% \\
Pass 2 pruning & 95.17\% & 95.61\% & 95.21\% \\
Pass 3 pruning & 94.92\% & 95.60\% & 95.00\% \\
Pass 4 pruning & 94.87\% & \textbf{95.66}\% &95.10\% \\
Pass 5 pruning & -- & 95.41\% & -- \\
Pass 6 pruning & -- & 95.47\% & -- \\
\hline

\end{tabular}
\end{center}
\end{table}

\section{Experiments}
\label{sec:experiments}

\subsection{Implementation Details}

{Our network is based on the Inflated ResNet50 as described in Section~\ref{sec:network}. We train our network for 20 epochs, with an adaptive learning rate initially set to 0.0001. We divide the learning rate by 10 when the validation loss has not improved for two consecutive epochs. We use a pruning ratio of 10\% at each pruning pass.}

{We preprocess all input videos with cropping and rescaling as described in Section~\ref{sec:preprocessing}. In addition, we flip all the video frames horizontally with a probability of 50\%. For both training and testing, we partition each video clip into eight equal segments and uniformly sample eight frames as the input frames, with the location of the first frame randomly chosen within the first segment. For training we also randomly crop a fixed $224 \times 224$ square from all input frames to feed into our network, but at test time we crop a square of the same size at the center of each frame. We also shuffle the training data at the beginning of each epoch. For testing, five sets of frames are sampled from the eight segments of each clip as input to the network, and the final classification probabilities are averaged similar to \cite{Baradel_2018}. $10\%$ of the test data is randomly chosen as the validation data.} 

\subsection{Network Architecture}
\label{sec:resnet}
{We perform experiments to search for the best Inflated ResNet architecture as shown in Table~\ref{tbl:Inflated_architecture_comparison}. The original ResNet has five variations: ResNet18, ResNet34, ResNet50, ResNet101 and ResNet152. We do experiments on the inflated version of the first four variations, as ResNet152 is unnecessarily big for our datasets.}

{We note that the Inflated ResNet34 has more parameters than the Inflated ResNet50, as they are made of different building blocks. We refer the readers to \cite{carreira2017quo} for more details of the inflation procedure. From Table~\ref{tbl:Inflated_architecture_comparison} we conclude that the Inflated ResNet50 is the best baseline model to use for our further experiments.}

\begin{table}
\begin{center}
\caption{Different Inflated ResNet Architectures on NTU 60.}
\label{tbl:Inflated_architecture_comparison}
\begin{tabular}{|c|c|c|c|}
\hline
Architecture & \#Parameters& Cross-Subject & Cross-View\\
\hline
ResNet18 & 33.7M & 93.83\% & 97.87\%\\
ResNet34 & 64.0M & 93.91\% & 97.69\%\\
ResNet50 & 46.8M & \textbf{95.25}\% & \textbf{98.34}\%\\
ResNet101 & 85.8M & 95.19\% & 98.09\%\\
\hline

\end{tabular}
\end{center}
\end{table}

\subsection{Hyperparameters}
\label{sec:hyperparameters}
{The number of superclasses for each level of the hierarchical classification is one important hyperparameter of our model. However, when the original number of action classes is big, such as 60 for NTU 60, there are too many combinations of these parameters for all the classification levels. We thus chose some representative combinations to test for NTU 60 and the results are shown in Table~\ref{tbl:super_class_comparison}. We have also performed experiments on different weighting schemes for the hierarchical loss function in Table~\ref{tbl:loss_weigths_comparison}, and different pruning ratios in Table~\ref{tbl:pruning_comparison}.}

\begin{table}
\begin{center}
\caption{Different superclass sizes for level 1-3 classification on the NTU 60 dataset.}
\label{tbl:super_class_comparison}
\begin{tabular}{|c|c|c|}
\hline
Superclasses & Cross-Subject & Cross-View\\
\hline
2/6/20 & \textbf{95.45}\% & \textbf{98.59}\% \\
3/10/30 & 95.25\% & 98.34\%\\
4/12/30 & 95.25\% & 98.38\%\\
5/15/30 & 95.20\% & 98.29\%\\
6/12/30 & 95.39\% & 98.28\%\\
10/20/30 & 95.04\% & 98.38\%\\
60/60/60 & 95.21\% & 98.37\%\\
\hline

\end{tabular}
\end{center}
\end{table}



\subsection{Comparisons}

{We evaluate our method and compare with other state-of-the-art methods on four commonly used datasets: NTU RGB+D 60  Dataset (NTU 60), NTU RGB+D 120 Dataset (NTU 120), Northwestern-UCLA Multiview Action 3D Dataset (N-UCLA), and UTD Multimodal Human Action Dataset (UTD-MHAD). For all these datasets, we crop the videos in a preprocessing stage as described in Section~\ref{sec:preprocessing}. In all experiments, we set the number of superclasses heuristically based on testings as described in Section~\ref{sec:hyperparameters}. More dataset-specific settings are given in the dataset descriptions next.}


\subsubsection{Datasets}
\label{sec:datasets}

{\textbf{NTU RGB+D 60 Dataset (NTU 60)} contains more than 56000 video clips~\cite{7780484}. 2D and 3D skeleton data, as well as depth, are also available. There are 60 action classes, and two evaluation benchmarks: cross-view and cross-subject. We choose 2, 6, 20 as the number of superclasses for levels 1-3 classification respectively, according to Table~\ref{tbl:super_class_comparison}. We use weights $(\frac{1}{8},\frac{1}{4},\frac{1}{2},1)$ for the hierarchical loss function, according to Table~\ref{tbl:loss_weigths_comparison}.}

\begin{table*}
\begin{center}
\caption{Comparison on NTU 60. -- indicates no results available.}
\label{tbl:NTU_result}
\begin{tabular}{|c|c|c|c|c|c|}
\hline
Method & Year & Pose Input & RGB Input & Cross-View & Cross-Subject \\
\hline
Glimpse Clouds {\cite{Baradel_2018}} & 2018 &  & \checkmark & 93.2\% & 86.6\%\\
FGCN {\cite{yang2020feedback}} & 2020 & \checkmark &  & 96.25\% & 90.22\%\\
MS-G3D Net {\cite{liu2020disentangling}} & 2020 & \checkmark &  & 96.2\%& 91.5\%\\
PoseMap {\cite{Liu2018Recognizing}}& 2018 & \checkmark & \checkmark & 95.26\% & 91.71\% \\
MMTM {\cite{joze2019mmtm}}& 2019 & \checkmark & \checkmark & -- & 91.99\% \\
Action Machine {\cite{zhu2018action}} & 2019 &  & \checkmark & 97.2\% & 94.3\%\\
PGCN {\cite{shi2019action}} & 2019 & \checkmark & \checkmark & -- & \textbf{96.4}\%\\
\hline
ours & 2020 & \checkmark & \checkmark & \textbf{98.79}\% & 95.66\% \\
\hline
\end{tabular}
\end{center}
\end{table*}

{\textbf{NTU RGB+D 120 Dataset (NTU 120)} adds 60 new action classes to the original NTU 60 dataset~\cite{Liu_2019}. It contains more than 114000 video clips in total and provides two benchmarks: cross-setup and cross-subject. As the number of action classes is doubled compared with that of NTU 60, we use 4,12,40 as the number of superclasses for levels 1-3 classification. We still use $(\frac{1}{8},\frac{1}{4},\frac{1}{2},1)$ for weighting the hierarchical loss function.}

\begin{table*}
\begin{center}
\caption{Comparison on NTU 120. * indicates results obtained from author-released code. -- indicates no results available. }
\label{tbl:NTU120_result}
\begin{tabular}{|c|c|c|c|c|c|}
\hline
Method & Year & Pose Input & RGB Input & Cross-Setup & Cross-Subject\\
\hline
Action Machine {\cite{zhu2018action}} & 2019 &  & \checkmark & -- & --\\
TSRJI {\cite{caetano2019skeleton}} & 2019 & \checkmark &  & 62.8\% & 67.9\%\\
PoseMap from Papers with Code~{\cite{PoseMap_paperswithcode}} & 2018 & \checkmark & \checkmark & 66.9\% & 64.6\% \\
SkeleMotion {\cite{Caetano_2019}} & 2019 & \checkmark &  & 66.9\% & 67.7\%\\
GVFE + AS-GCN with DH-TCN {\cite{papadopoulos2019vertex}} & 2019 & \checkmark &  & 79.8\% & 78.3\% \\
Glimpse Clouds  {\cite{Baradel_2018}} & 2018 &  & \checkmark & 83.84\%* & 83.52\%*\\
FGCN {\cite{yang2020feedback}} & 2020 & \checkmark & & 87.4\% & 85.4\%\\
MS-G3D Net {\cite{liu2020disentangling}} & 2020 & \checkmark & & 88.4\%& 86.9\%\\
\hline
ours & 2020 & \checkmark & \checkmark & \textbf{94.54}\% & \textbf{93.69}\%  \\

\hline
\end{tabular}
\end{center}
\end{table*}

{\textbf{Northwestern-UCLA Multiview Action 3D (N-UCLA)} contains 1494 video sequences, together with depth and 3D skeleton data~\cite{DBLP:journals/corr/WangNXWZ14}. Each action is recorded simultaneously with three Kinect cameras. We convert the 3D skeleton data into 2D skeletons using the projection method described in Section~\ref{sec:preprocessing}. We use three view-based benchmarks where each view is used for testing and the other two for training. There are 10 different actions in this dataset. We thus choose 2 and 5 as the number of superclasses for the level-2 and level-3 classifiers, respectively. We do not need the level-1 classifier for hierarchical classification anymore. We use weights $(\frac{1}{16},\frac{1}{4},1)$ for the hierarchical loss function. As the size of this dataset is small, we use the pre-trained network on NTU 60 cross-subject benchmark to initialize the network training on N-UCLA.} 

\begin{table*}
\begin{center}
\caption{Comparison on N-UCLA. -- indicates no results available. The Pre-trained column indicates if the model was pre-trained on ImageNet and/or a bigger human action dataset. }
\label{tbl:NUCLA_result}
\begin{tabular}{|c|c|c|c|c|c|c|c|c|}
\hline
Method & Year & Pre-trained & Pose Input & RGB Input & View1 & View2 & View3 & Average\\
\hline
PoseMap {\cite{Liu2018Recognizing}} & 2018 & \checkmark &\checkmark & \checkmark & -- & -- & -- & --\\
Ensemble TS-LSTM {\cite{Lee2017enesmble}} & 2017 & & \checkmark &  & -- & -- & 89.22\% & --\\
EleAtt-GRU(aug.) {\cite{zhang2018adding}} & 2018 &  \checkmark & \checkmark &  & -- & -- & 90.7\% & --\\
Enhanced Viz. {\cite{articleLiu}} & 2017 & \checkmark&\checkmark &  & -- & -- & 92.61\%& --\\
Glimpse Clouds {\cite{Baradel_2018}} & 2018 & \checkmark &  & \checkmark & 83.4\% & 89.5\% & 90.1\% & 87.6\%\\
FGCN {\cite{yang2020feedback}} & 2020 & & \checkmark &  &  -- & -- & 95.3\% & --\\
Action Machine {\cite{zhu2018action}} & 2019 &  \checkmark &  & \checkmark & 88.3\% & \textbf{92.2}\% & 96.5\% & 92.3\%\\
\hline
ours & 2020 & \checkmark &  \checkmark & \checkmark & \textbf{91.10}\% & 91.95\% & \textbf{98.92}\% & \textbf{93.99}\%\\

\hline
\end{tabular}
\end{center}
\end{table*}

{\textbf{UTD Multimodal Human Action Dataset (UTD-MHAD)} contains 861 video sequences, together with depth, 3D skeleton, and wearable inertial sensor data~\cite{7350781}. It provides one cross-subject evaluation benchmark. Similar as for the N-UCLA dataset, we convert the 3D skeleton data into 2D skeletons by projection and use the pre-trained network on NTU 60 for initialization. There are 27 action classes so we choose 3 and 9 as the number of superclasses for the level-2 and level-3 classifiers, respectively. We also use weights $(\frac{1}{16},\frac{1}{4},1)$ for the hierarchical loss function, similar as for the N-UCLA dataset.}

\begin{table*}
\begin{center}
\caption{Comparison on UTD-MHAD. * indicates results obtained from author-released code. -- indicates no results available. The Pre-trained column indicates if the model was pre-trained on ImageNet and/or a bigger human action dataset.}
\label{tbl:UTD_result}
\begin{tabular}{|c|c|c|c|c|c|c|c|}
\hline
Method & Year & Pre-trained & Pose Input & RGB Input & Cross-Subject\\
\hline
Glimpse Clouds {\cite{Baradel_2018}} & 2018 & \checkmark &   & \checkmark & 84.19\%*\\
JTM {\cite{Wang_2016}} & 2016 & \checkmark & \checkmark &  & 85.81\% \\
Optical Spectra {\cite{Hou2018Skeleton}} & 2018 & \checkmark & \checkmark &  & 86.97\% \\
JDM {\cite{Li2017Joint}} & 2017 & \checkmark & \checkmark &  & 88.10\% \\
Action Machine Archived Version{\cite{zhu2018action_arxiv}} & 2019 & \checkmark  &  & \checkmark & 92.5\% \\
PoseMap {\cite{Liu2018Recognizing}} & 2018 & \checkmark & \checkmark & \checkmark & \textbf{94.51}\% \\
\hline
ours & 2020 & \checkmark & \checkmark & \checkmark & 91.63\% \\

\hline
\end{tabular}
\end{center}
\end{table*}



\subsubsection{Performance}
{Tables~\ref{tbl:NTU_result}-\ref{tbl:UTD_result} show the comparison results. Our method scores the highest or close to the highest for all four datasets. We report the accuracy for prior work by either directly taking numbers reported in the original paper, or running author-released code if relevant performance was not reported in the original papers. For fair comparisons, we check if a method uses RGB input or pose input or both. We also mark if a method uses pre-trained models on ImageNet and/or a bigger human action dataset to initialize the network for training on small datasets.}

\section{Ablation Study}

{We detail the performance gains of the video cropping preprocess, the hierarchical classification, and the multipass network pruning components with ablation studies shown in Tables \ref{tbl:NTU_60_ab}-\ref{tbl:UTD_ab}. The hierarchical classification is proved to be beneficial in all cases after cropping. Without cropping, the hierarchical classification alone is also able to gain performance as shown in the third row of Table~\ref{tbl:NTU_120_ab} for NTU 120.} 

{We continue the network pruning iterations until we observe lowered performance from the best one achieved so far in two consecutive steps. The network pruning is helpful in most cases. 
For small datasets, pruning layer by layer outperforms pruning all layers together. For larger datasets such as the NTU datasets, the two pruning methods perform comparably.}

{The cropping preprocess yields large performance gains in most cases, but not for the UTD-MHAD dataset, which was captured in one single environment with actors located in the frame centers already.}

\begin{table*}
\begin{center}
\caption{Ablation Study on NTU 60 dataset. The baseline model refers to the Inflated ResNet50 network trained without hierarchical loss using the original videos. The full model refers to the baseline network trained with the hierarchical loss using cropped video input.}
\label{tbl:NTU_60_ab}
\begin{tabular}{|c|c|c|c|c|}
\hline
 & \multicolumn{2}{c|}{Pruning Altogether} & \multicolumn{2}{c|}{Pruning by Layers}\\
\hline
Method & Cross-Subject & Cross-View & Cross-Subject & Cross-View \\
\hline
baseline & 89.13\% & 94.39\% & 89.13\% & 94.39\% \\
baseline + cropping & 95.29\% & 98.34\% & 95.29\% & 98.34\% \\
full model & 95.45\% & 98.59\% & 95.45\% & 98.59\% \\
full model + 1-pass pruning & 95.52\% & 98.61\% & 95.37\% & 98.66\%\\
full model + 2-pass pruning & 95.61\% & 98.60\% & 95.50\% & 98.63\%\\
full model + 3-pass pruning & 95.60\% & \textbf{98.65}\% & \textbf{95.64}\% & 98.74\%\\
full model + 4-pass pruning & \textbf{95.66}\% & 98.40\% & 95.14\% & \textbf{98.79}\%\\
full model + 5-pass pruning & 95.41\% & 98.43\% & 95.39\% & 98.56\%\\
full model + 6-pass pruning & 95.47\% & -- & -- & 98.62\% \\
\hline
\end{tabular}
\end{center}
\end{table*}


\begin{table*}
\begin{center}
\caption{Ablation Study on NTU 120 dataset.}
\label{tbl:NTU_120_ab}
\begin{tabular}{|c|c|c|c|c|}
\hline
 & \multicolumn{2}{c|}{Pruning Altogether} & \multicolumn{2}{c|}{Pruning by Layers}\\
\hline
Method & Cross-Subject & Cross-Setup & Cross-Subject & Cross-Setup \\
\hline
baseline & 84.64\% & 86.20\% & 84.64\% & 86.20\% \\
baseline + cropping & 92.48\% & 93.99\% & 92.48\% & 93.99\% \\
baseline + hierarchical loss & 86.17\% & 86.98\% & 86.17\% & 86.98\% \\
full model & 92.95\% & 94.25\% & 92.95\% & 94.25\% \\
full model + 1-pass pruning & 93.48\% & 94.12\% & 93.19\% & \textbf{94.54}\%\\
full model + 2-pass pruning & 93.55\% & 94.31\% & \textbf{93.56}\% & 94.48\% \\
full model + 3-pass pruning & \textbf{93.69}\% & \textbf{94.42}\% & 93.43\% & 94.48\% \\
full model + 4-pass pruning & 93.64\% & 94.39\% & 93.38\% & --\\
full model + 5-pass pruning & 93.47\% & 94.19\% & -- & --\\
\hline
\end{tabular}
\end{center}
\end{table*}

\begin{table*}
\begin{center}
\caption{Ablation Study on N-UCLA dataset. The ``Viewn'' columns indicate the camera view used for testing and the other two views are used for training.}
\label{tbl:NUCLA_ab}
\begin{tabular}{|c|c|c|c|c|c|c|}
\hline
& \multicolumn{3}{c|}{Pruning Altogether} & \multicolumn{3}{c|}{Pruning by Layers}\\
\hline
Method & View1 & View2 & View3 & View1 & View2 & View3 \\
\hline

baseline & 77.76\% & 70.73\% & 97.00\% & 77.76\% & 70.73\% & 97.00\%  \\
baseline + cropping & 88.78\% & 89.00\% & 98.50\% & 88.78\% & 89.00\% & 98.50\% \\
full model & 88.78\% & \textbf{89.98}\% & \textbf{98.92}\% & 88.78\% & 89.98\% & \textbf{98.92}\% \\
full model + 1-pass pruning & 89.75\% & 86.64\% & 98.28\% & 88.39\% & 88.39\% & 98.71\%\\
full model + 2-pass pruning & \textbf{90.14}\% & 89.00\% & 98.28\% & \textbf{91.10}\% & 88.39\% & 98.07\%\\
full model + 3-pass pruning & 71.95\% & -- & 84.36\% & 86.85\% & 91.36\% & --\\
full model + 4-pass pruning & 40.62\% & -- & -- & 85.88\% & 90.37\% & -- \\
full model + 5-pass pruning & -- & -- & -- & -- & \textbf{91.95}\% & -- \\
full model + 6-pass pruning & -- & -- & -- & -- & 90.37\% & -- \\
full model + 7-pass pruning & -- & -- & -- & -- & 88.21\% & -- \\

\hline
\end{tabular}
\end{center}
\end{table*}

\begin{table*}
\begin{center}
\caption{Ablation Study on UTD-MHAD dataset.}
\label{tbl:UTD_ab}
\begin{tabular}{|c|c|c|}
\hline
 & Pruning Altogether & Pruning by Layers\\
\hline
Method & Cross-Subject & Cross-Subject \\
\hline
baseline & 84.88 \% & 84.88 \% \\
baseline + cropping & 82.55\% & 82.55\% \\
full model& \textbf{86.05}\% & 86.05\% \\
full model + 1-pass pruning & 84.65\% & 87.44\% \\
full model + 2-pass pruning & 52.23\% & 86.51\% \\
full model + 3-pass pruning & -- & 90.46\% \\
full model + 4-pass pruning & -- & \textbf{91.63}\%\\
full model + 5-pass pruning & -- & 90.23\% \\
full model + 6-pass pruning & -- & 90.93\% \\

\hline
\end{tabular}
\end{center}
\end{table*}
\section{Conclusions}
{We have augmented the Inflated ResNet50 architecture with hierarchical classification, iterative network pruning, and skeleton-based cropping. These components are simple to implement and effective in improving the classification accuracy for human action datasets captured in lab environments. Our work has set up a new baseline for the NTU 120 dataset, which is the largest dataset of its kind.} 

\subsection{Limitations and Future Work}
{Our choices of hyperparameters, such as the number of superclasses for each classification level, number of iterations for the pruning, and weights in the hierarchical loss function, are all set by either heuristics and/or manual searching. There is no guarantee that these parameters are optimal. Automatic search algorithms that are not computationally prohibitive are desirable.} 

{We have used the skeleton data only for video preprocessing in this work. GCN-based skeleton classification algorithms, however, can be easily incorporated to further boost the performance of our system. We could simply fuse the classification results of the two algorithms, or implement an integrated two-stream system from both the video stream and the skeleton stream.}

{We have mainly focused on NTU RGB+D type of datasets that were captured in lab settings together with 2D/3D skeletons of reasonable quality. For future work, we would like to incorporate outdoor datasets or datasets obtained from the Internet, such as the Kinetics dataset~\cite{kay2017kinetics}. Such datasets are usually more heterogeneous and of much lower quality. 2D skeletons extracted by pose estimation algorithms such as OpenPose~\cite{OpenPose} are also less trustworthy. Large-scale pretraining using super large neural network models is likely needed as suggested by \cite{duan2020omnisourced,Ghadiyaram_2019,Feichtenhofer_2019}.}

\bibliographystyle{unsrt}  
\bibliography{hAction}

\begin{thebibliography}{10}

\bibitem{caetano2019skeleton}
Carlos Caetano, Francois Bremond, and William~Robson Schwartz.
\newblock Skeleton image representation for 3d action recognition based on tree
  structure and reference joints.
\newblock {\em SIBGRAPI Conference on Graphics, Patterns and Images}, 2019.

\bibitem{Caetano_2019}
Carlos Caetano, Jessica Sena, Franeois Bremond, Jefersson~A. Dos~Santos, and
  William~Robson Schwartz.
\newblock Skelemotion: A new representation of skeleton joint sequences based
  on motion information for 3d action recognition.
\newblock {\em IEEE International Conference on Advanced Video and Signal Based
  Surveillance}, 2019.

\bibitem{articleLiu}
Mengyuan Liu, Hong Liu, and Chen Chen.
\newblock Enhanced skeleton visualization for view invariant human action
  recognition.
\newblock {\em Pattern Recognition}, 03 2017.

\bibitem{Wang_2016}
Pichao Wang, Zhaoyang Li, Yonghong Hou, and Wanqing Li.
\newblock Action recognition based on joint trajectory maps using convolutional
  neural networks.
\newblock {\em ACM International Conference on Multimedia}, 2016.

\bibitem{Li2017Joint}
C.~{Li}, Y.~{Hou}, P.~{Wang}, and W.~{Li}.
\newblock Joint distance maps based action recognition with convolutional
  neural networks.
\newblock {\em IEEE Signal Processing Letters}, 24(5):624--628, 2017.

\bibitem{Hou2018Skeleton}
Y.~{Hou}, Z.~{Li}, P.~{Wang}, and W.~{Li}.
\newblock Skeleton optical spectra-based action recognition using convolutional
  neural networks.
\newblock {\em IEEE Transactions on Circuits and Systems for Video Technology},
  28(3):807--811, 2018.

\bibitem{Lee2017enesmble}
I.~{Lee}, D.~{Kim}, S.~{Kang}, and S.~{Lee}.
\newblock Ensemble deep learning for skeleton-based action recognition using
  temporal sliding lstm networks.
\newblock In {\em IEEE International Conference on Computer Vision}, pages
  1012--1020, 2017.

\bibitem{zhang2018adding}
Pengfei Zhang, Jianru Xue, Cuiling Lan, Wenjun Zeng, Zhanning Gao, and Nanning
  Zheng.
\newblock Adding attentiveness to the neurons in recurrent neural networks.
\newblock In {\em Proceedings of the European Conference on Computer Vision},
  pages 135--151, 2018.

\bibitem{liu2020disentangling}
Ziyu Liu, Hongwen Zhang, Zhenghao Chen, Zhiyong Wang, and Wanli Ouyang.
\newblock Disentangling and unifying graph convolutions for skeleton-based
  action recognition, 2020.

\bibitem{Shi_2019_Skeleton}
L.~{Shi}, Y.~{Zhang}, J.~{Cheng}, and H.~{Lu}.
\newblock Skeleton-based action recognition with directed graph neural
  networks.
\newblock In {\em IEEE Conference on Computer Vision and Pattern Recognition},
  pages 7904--7913, 2019.

\bibitem{shi2018twostream}
Lei Shi, Yifan Zhang, Jian Cheng, and Hanqing Lu.
\newblock Two-stream adaptive graph convolutional networks for skeleton-based
  action recognition.
\newblock In {\em The IEEE Conference on Computer Vision and Pattern
  Recognition}, 2019.

\bibitem{yang2020feedback}
Hao Yang, Dan Yan, Li~Zhang, Dong Li, YunDa Sun, ShaoDi You, and Stephen~J.
  Maybank.
\newblock Feedback graph convolutional network for skeleton-based action
  recognition, 2020.

\bibitem{papadopoulos2019vertex}
Konstantinos Papadopoulos, Enjie Ghorbel, Djamila Aouada, and Björn Ottersten.
\newblock Vertex feature encoding and hierarchical temporal modeling in a
  spatial-temporal graph convolutional network for action recognition, 2019.

\bibitem{carreira2017quo}
Joao Carreira and Andrew Zisserman.
\newblock Quo vadis, action recognition? a new model and the kinetics dataset.
\newblock {\em IEEE Conference on Computer Vision and Pattern Recognition},
  2017.

\bibitem{joze2019mmtm}
Hamid Reza~Vaezi Joze, Amirreza Shaban, Michael~L. Iuzzolino, and Kazuhito
  Koishida.
\newblock Mmtm: Multimodal transfer module for cnn fusion, 2019.

\bibitem{Baradel_2018}
Fabien Baradel, Christian Wolf, Julien Mille, and Graham~W. Taylor.
\newblock Glimpse clouds: Human activity recognition from unstructured feature
  points.
\newblock {\em IEEE Conference on Computer Vision and Pattern Recognition},
  2018.

\bibitem{zhu2018action}
Jiagang Zhu, Wei Zou, Zhu Zheng, Liang Xu, and Guan Huang.
\newblock Action machine: Toward person-centric action recognition in videos.
\newblock {\em IEEE Signal Processing Letters}, PP, 2019.

\bibitem{shi2019action}
Lei Shi, Yifan Zhang, Jian Cheng, and Han-Qing Lu.
\newblock Action recognition via pose-based graph convolutional networks with
  intermediate dense supervision.
\newblock {\em ArXiv}, abs/1911.12509, 2019.

\bibitem{Liu2018Recognizing}
M.~{Liu} and J.~{Yuan}.
\newblock Recognizing human actions as the evolution of pose estimation maps.
\newblock In {\em IEEE Conference on Computer Vision and Pattern Recognition},
  pages 1159--1168, 2018.

\bibitem{Ballin_3DFlowEstimation}
Gioia Ballin, Matteo Munaro, and Emanuele Menegatti.
\newblock Human action recognition from rgb-d frames based on real-time 3d
  optical flow estimation.
\newblock In Antonio Chella, Roberto Pirrone, Rosario Sorbello, and
  Kamilla~R{\'u}n J{\'o}hannsd{\'o}ttir, editors, {\em Biologically Inspired
  Cognitive Architectures 2012}, pages 65--74. Springer Berlin Heidelberg,
  2013.

\bibitem{Simonyan_two-stream}
Karen Simonyan and Andrew Zisserman.
\newblock Two-stream convolutional networks for action recognition in videos.
\newblock In {\em Proceedings of the International Conference on Neural
  Information Processing Systems - Volume 1}, page 568–576. MIT Press, 2014.

\bibitem{Perez_Rua_2019}
Juan-Manuel Perez-Rua, Valentin Vielzeuf, Stephane Pateux, Moez Baccouche, and
  Frederic Jurie.
\newblock Mfas: Multimodal fusion architecture search.
\newblock {\em IEEE Conference on Computer Vision and Pattern Recognition},
  2019.

\bibitem{Kawahara2016MultiresolutionTractCW}
Jeremy Kawahara and Ghassan Hamarneh.
\newblock Multi-resolution-tract cnn with hybrid pretrained and skin-lesion
  trained layers.
\newblock In Li~Wang, Ehsan Adeli, Qian Wang, Yinghuan Shi, and Heung-Il Suk,
  editors, {\em Machine Learning in Medical Imaging}, pages 164--171. Springer
  International Publishing, 2016.

\bibitem{Wu_2019}
Cinna Wu, Mark Tygert, and Yann LeCun.
\newblock A hierarchical loss and its problems when classifying
  non-hierarchically.
\newblock {\em PLOS ONE}, 14(12), 2019.

\bibitem{Szegedy_2015_CVPR}
Christian Szegedy, Wei Liu, Yangqing Jia, Pierre Sermanet, Scott Reed, Dragomir
  Anguelov, Dumitru Erhan, Vincent Vanhoucke, and Andrew Rabinovich.
\newblock Going deeper with convolutions.
\newblock In {\em The IEEE Conference on Computer Vision and Pattern
  Recognition}, June 2015.

\bibitem{huang2018multiscale}
Gao Huang, Danlu Chen, Tianhong Li, Felix Wu, Laurens van~der Maaten, and
  Kilian Weinberger.
\newblock Multi-scale dense networks for resource efficient image
  classification.
\newblock In {\em International Conference on Learning Representations}, 2018.

\bibitem{frankle2018lottery}
Jonathan Frankle and Michael Carbin.
\newblock The lottery ticket hypothesis: Finding sparse, trainable neural
  networks.
\newblock In {\em International Conference on Learning Representations}, 2019.

\bibitem{liu2018rethinking}
Zhuang Liu, Mingjie Sun, Tinghui Zhou, Gao Huang, and Trevor Darrell.
\newblock Rethinking the value of network pruning.
\newblock In {\em International Conference on Learning Representations}, 2019.

\bibitem{li2016pruning}
Hao Li, Asim Kadav, Igor Durdanovic, Hanan Samet, and Hans~Peter Graf.
\newblock Pruning filters for efficient convnets.
\newblock In {\em International Conference on Learning Representations}, 2017.

\bibitem{OpenPose}
Zhe Cao, Tomas Simon, Shih-En Wei, and Yaser Sheikh.
\newblock Realtime multi-person 2d pose estimation using part affinity fields.
\newblock {\em IEEE Conference on Computer Vision and Pattern Recognition},
  2017.

\bibitem{He_2016}
Kaiming He, Xiangyu Zhang, Shaoqing Ren, and Jian Sun.
\newblock Deep residual learning for image recognition.
\newblock {\em IEEE Conference on Computer Vision and Pattern Recognition},
  2016.

\bibitem{7780484}
A.~{Shahroudy}, J.~{Liu}, T.~{Ng}, and G.~{Wang}.
\newblock Ntu rgb+d: A large scale dataset for 3d human activity analysis.
\newblock In {\em IEEE Conference on Computer Vision and Pattern Recognition},
  pages 1010--1019, 2016.

\bibitem{Liu_2019}
Jun Liu, Amir Shahroudy, Mauricio~Lisboa Perez, Gang Wang, Ling-Yu Duan, and
  Alex Kot~Chichung.
\newblock Ntu rgb+d 120: A large-scale benchmark for 3d human activity
  understanding.
\newblock {\em IEEE Transactions on Pattern Analysis and Machine Intelligence},
  2019.

\bibitem{PoseMap_paperswithcode}
M.~{Liu} and J.~{Yuan}.
\newblock Recognizing human actions as the evolution of pose estimation maps.
\newblock
  \url{https://paperswithcode.com/paper/recognizing-human-actions-as-the-evolution-of}.
\newblock Accessed: 2020-05-12.

\bibitem{DBLP:journals/corr/WangNXWZ14}
J.~{Wang}, X.~{Nie}, Y.~{Xia}, Y.~{Wu}, and S.~{Zhu}.
\newblock Cross-view action modeling, learning, and recognition.
\newblock {\em IEEE Conference on Computer Vision and Pattern Recognition},
  pages 2649--2656, 2014.

\bibitem{7350781}
C.~{Chen}, R.~{Jafari}, and N.~{Kehtarnavaz}.
\newblock Utd-mhad: A multimodal dataset for human action recognition utilizing
  a depth camera and a wearable inertial sensor.
\newblock In {\em IEEE International Conference on Image Processing}, pages
  168--172, 2015.

\bibitem{zhu2018action_arxiv}
Jiagang Zhu, Wei Zou, Liang Xu, Yiming Hu, Zheng Zhu, Manyu Chang, Junjie
  Huang, Guan Huang, and Dalong Du.
\newblock Action machine: Rethinking action recognition in trimmed videos,
  2018.

\bibitem{kay2017kinetics}
Will Kay, Joao Carreira, Karen Simonyan, Brian Zhang, Chloe Hillier, Sudheendra
  Vijayanarasimhan, Fabio Viola, Tim Green, Trevor Back, Paul Natsev, Mustafa
  Suleyman, and Andrew Zisserman.
\newblock The kinetics human action video dataset, 2017.

\bibitem{duan2020omnisourced}
Haodong Duan, Yue Zhao, Yuanjun Xiong, Wentao Liu, and Dahua Lin.
\newblock Omni-sourced webly-supervised learning for video recognition, 2020.

\bibitem{Ghadiyaram_2019}
Deepti Ghadiyaram, Du~Tran, and Dhruv Mahajan.
\newblock Large-scale weakly-supervised pre-training for video action
  recognition.
\newblock {\em IEEE Conference on Computer Vision and Pattern Recognition},
  2019.

\bibitem{Feichtenhofer_2019}
Christoph Feichtenhofer, Haoqi Fan, Jitendra Malik, and Kaiming He.
\newblock Slowfast networks for video recognition.
\newblock {\em IEEE International Conference on Computer Vision}, 2019.

\end{thebibliography}

\end{document}